\documentclass{article}
\usepackage[utf8]{inputenc}

\usepackage{PRIMEarxiv}

\usepackage[utf8]{inputenc} 
\usepackage[T1]{fontenc}    
\usepackage{hyperref}       
\usepackage{url}            
\usepackage{booktabs}       
\usepackage{amsfonts}       
\usepackage{nicefrac}       
\usepackage{microtype}      
\usepackage{fancyhdr}       
\usepackage{graphicx}       
\graphicspath{{media/}}     
\usepackage{adjustbox}
\usepackage{booktabs} 
\usepackage{kotex}
\usepackage{listings}
\usepackage{amsmath}
\usepackage{caption}
\usepackage{subcaption}
\usepackage{tikz}
\usetikzlibrary{arrows.meta,positioning}

\title{SLM-Based Agentic AI with P–C–G: Optimized for Korean Tool Use
}

\author{
  Changhyun Jeon, Jinhee Park, Jungwoo Choi, Keonwoo Kim, Jisu Kim, Minji Hong \\
  HANCOM / Seongnam, South Korea \\
  \texttt{\{jeonch, jhpark1, jungwoo.choi, keonwoo.kim, jisu.kim, minji.hong\}@hancom.com} \\
}

\begin{document}
\maketitle
\setcounter{footnote}{0}

\begin{abstract}
We propose a small-scale language model (SLM) based agent architecture, \textbf{Planner–Caller–Generator (P–C–G)}, optimized for Korean tool use. P–C–G separates planning, calling, and generation by role: the \textit{Planner} produces an initial batch plan with \emph{limited on-demand replanning}; the \textit{Caller} returns a normalized call object after joint schema–value validation; and the \textit{Generator} integrates tool outputs to produce the final answer. We apply a \emph{Korean-first value policy} to reduce execution failures caused by frequent Korean-to-English code switching in Korean settings. Evaluation assumes Korean queries and Korean tool/parameter specifications; it covers single-chain, multi-chain, missing-parameters, and missing-functions scenarios, and is conducted via an \emph{LLM-as-a-Judge} protocol averaged over five runs under a unified I/O interface. Results show that P–C–G delivers competitive tool-use accuracy and end-to-end quality while reducing tokens and maintaining acceptable latency, indicating that role-specialized SLMs are a cost-effective alternative for Korean tool-use agents.
\end{abstract}


\section{Introduction}
Recently, Large Language Models (LLMs) have attracted significant attention by achieving performance approaching human level across a wide range of natural language understanding and generation tasks. However, LLMs with tens to hundreds of billions of parameters require substantial computational resources and cost~\cite{achiam2023gpt, carlini2021extracting}. To mitigate these costs, small-scale language models (SLMs) with greatly reduced parameter counts have emerged, and recent studies indicate that such smaller models can, in certain settings, attain performance comparable to LLM~\cite{dubey2024llama, yang2025qwen3}.

However, because these language models fundamentally operate via next-token prediction based on pretraining data, they struggle to incorporate up-to-date or rapidly changing information. They also continue to exhibit hallucinations—generating content that is not factually grounded. While continual training with new data can mitigate these issues, it incurs substantial additional cost. Consequently, there is growing interest in Agentic AI systems that combine language models with tool use mechanisms—such as retrieval-augmented generation (RAG) and API/tool calls—to provide up-to-date information cost-effectively while addressing core limitations of standalone language models.

Agentic AI systems enable a language model to orchestrate external tools to carry out complex tasks, thereby compensating for LLM limitations in accessing current information, producing action-grounded responses, and performing multi-step reasoning. By integrating functions such as search, calculation, and API calls, agentic AI systems can solve various real-world problems and have advantages in terms of automation and scalability. However, most existing Agentic AI systems still rely heavily on LLMs, which entails high computational and monetary cost, as well as latency and privacy risks. The complexity of these systems also increases the potential for errors between modules, making overall robustness a key challenge.

To address these limitations, we propose \textbf{an Agentic AI system based on the Planner–Caller–Generator (P–C–G) architecture} that combines the efficiency of SLMs with the strengths of LLM-style agent techniques. Each SLM is specialized for a single role (task), with the goal of achieving LLM-comparable performance under tighter parameter size. Additionally, noting the relative lack of prior work that systematically evaluates tool use in Korean settings, our work focuses on \emph{reliable tool use on Korean data}.

This study investigates three questions: \
(1) Can an agent architecture that separates SLM roles (P-C-G) achieve a practically usable level of agent performance? \
(2) Does the planning approach we propose (\emph{initial planning + limited on-demand replanning}) actually reduce token usage and delays? \
(3) Does the system remain robust under multi-chain tool use and constraint scenarios (Missing Parameters/Functions)? \

To answer these questions, we present the P–C–G design with an \emph{initial planning + limited on-demand replanning} strategy, and evaluate Korean tool use across single-chain and multi-chain while explicitly including "missing" cases. We also enforce matched input/output formats and a common tool use interface for non-P–C–G baselines to ensure fair comparison, and demonstrate that that an 8B SLM can achieve meaningful performance on multi-chain tasks and constraint awareness.

Our scope and assumptions are as follows: we target tools with deterministic API responses, and separate the three roles of planning, tool use, and response generation. We assume inputs are de-identified with respect to personal data; security threats such as prompt injection are discussed as limitations and future work. The proposed design reduces unnecessary planning iterations to lower token usage and average latency, and is engineered to operate reliably on lightweight hardware.

For evaluation, we construct a Korean tool-use dataset (\emph{Single-chain, Multi-chain, Missing Parameters, Missing Functions}). We validate the performance using an \emph{LLM-as-a-Judge} protocol averaged over five runs, and compare it with publicly available large-scale models under identical conditions. This demonstrates that the SLM-based P–C–G architecture can serve as a viable alternative for practical agentic AI in the Korean environment.

\section{Related Works}
\subsection{LLM-based Agentic AI Systems}
With the rapid improvement of LLM capabilities, research on agentic AI that integrates external tools to perform complex tasks has accelerated~\cite{wang2024survey}. Early approaches typically relied on a single LLM to handle both reasoning and tool use. ReAct~\cite{yao2023react} interleaves \emph{reasoning} and \emph{action} to elicit tool calls and feed observations back into subsequent reasoning, thereby improving complex task solving. Toolformer~\cite{schick2023toolformer} expands tool-use ability by having an LLM annotate and learn where and how to invoke APIs during training. OpenAI's function calling\footnote{\url{https://platform.openai.com/docs/guides/function-calling?api-mode=responses}}
 provides JSON-based, explicit function invocation and has catalyzed ecosystems of agent frameworks such as LangChain, AutoGPT\footnote{\url{https://github.com/Significant-Gravitas/AutoGPT}}
, and BabyAGI\footnote{\url{https://github.com/yoheinakajima/babyagi}}
. These developments suggest an evolution from LLMs as mere text generators to central executors that coordinate external systems. Nevertheless, single-LLM–centric designs still face limitations, including token and latency overheads, hallucination, and security/privacy risks.

\subsection{Planning-Centric Agentic AI}
There has been a continuous effort to separate \emph{planning} from \emph{execution} to structure task solving more systematically. ReAct increases autonomy through a plan–act–observe loop, but the repetitive iterations incur token and cost overheads. ReWoo~\cite{xu2023rewoo} decouples planning from execution/observation and performs only the necessary invocations stepwise, reducing redundant token usage. Going further, LLMCompiler~\cite{sasabuchi2025plan, kim2024llm} decomposes a request into units composable in parallel or in series, and automatically synthesizes and schedules tool calls. These approaches improve efficiency through (i) explicit plans, (ii) caching of intermediate artifacts, and (iii) increasing parallelizability; however, operational challenges remain—detecting plan–state mismatch, recovering from errors, and verifying tool selection/argument correctness. In this paper, we propose an architecture that balances cost and stability by adopting a flow control that re-plans when necessary after initial batch planning, and by fixing the separation of roles between execution (Caller) and result integration (Generator).

\subsection{SLM-Based Agentic AI Systems}
Lightweight agent designs that employ SLMs have emerged as an alternative to LLM-centric agents, mitigating compute cost, latency, and privacy constraints while maintaining practical utility. Shen et al.~\cite{shen2024small} show that decomposing agent roles across multiple SLMs can compensate for weaknesses in tool use, and NVIDIA’s recent work~\cite{belcak2025small} argues that SLMs are intrinsically well-suited for agentic settings with respect to linguistic capacity, operational efficiency, and cost. This line of research shares a 'divide-and-specialize' design philosophy, emphasizing (i) role separation, (ii) standardized inter-module interfaces, and (iii) explicit control over tool invocation.

\subsection{Tool Use in the Korean Context}
Although tool-use benchmarks have rapidly advanced, most public datasets and evaluation protocols remain English-centric. API-Bank targets large sets of executable APIs but explicitly restricts itself to English~\cite{li2023apibankcomprehensivebenchmarktoolaugmented}; ToolBench\footnote{\url{https://github.com/OpenBMB/ToolBench}} and ToolLLM~\cite{qin2023toolllmfacilitatinglargelanguage} likewise build their tooling and specifications around English resources. In the Korean domain, widely used NLU benchmarks such as KLUE and KoBEST~\cite{DBLP:journals/corr/abs-2105-09680, jang-etal-2022-kobest} provide valuable coverage of understanding tasks, yet \emph{chained tool use} remains comparatively under-explored. FunctionChat-Bench~\cite{lee2024functionchatbenchcomprehensiveevaluationlanguage} moves in the direction of Korean tool-use evaluation, but focuses on benchmarking methodology rather than architectural designs for cost and stability.

This language mismatch has direct operational implications in Korean services. Values extracted from Korean user inputs are frequently used as arguments for search, reservation, or government-information tools. Unintended Korean-to-English switching in value fields can distort queries for Korean databases or alter user-facing outputs, leading to execution failures and degraded user experience. In addition, the lack of standardized Korean specifications for tool schemas (e.g., required fields, enumerations, admissible ranges) increases the risk of argument-format errors when models must infer or complete missing details. Robustness under multi-step dependencies---\emph{multi-chain} settings where outputs from one tool must be passed, possibly transformed, into the next---is especially sensitive to these issues.

Therefore, two main requirements stand out for the evaluation of Korean-centric tool-use. First, evaluation should \emph{assume Korean inputs and Korean tool/parameter specifications}, making value-handling policies explicit and penalizing unintended language switching that breaks semantics against Korean back-ends. For example, this includes cases such as inadvertently converting Korean personal or place names into English. Second, protocols should cover not only single-step calls but also \emph{multi-chain} dependencies and \emph{constraint} scenarios: \emph{Missing Parameters} (recognize absent required fields without hallucination) and \emph{Missing Functions} (detect that a necessary tool is not available and recover gracefully).

\paragraph{Positioning in the Korean context}
Motivated by the above gap, we adopt an SLM-based, role-separated \emph{Planner–Caller–Generator (P–C–G)} architecture tailored to reliable tool use in Korean. We enforce a \emph{Korean-first value policy} and apply \emph{schema/value co-validation} in the Caller to reduce execution failures. The Korean-first value policy keeps value fields in tool calls in Korean by default and allows conversion to another language (e.g., English) or codes only when the provided schema explicitly does not support Korean, following a narrowly scoped whitelist; this ensures language consistency between user-facing content and backend queries, with exceptions limited to schema-defined fields such as enumerations, regex/format constraints, or English-only fields. Reliability is evaluated under \emph{single-/multi-chain} tool use and \emph{constraint} scenarios (Missing Parameters/Functions) with a unified I/O interface and a five-run \emph{LLM-as-a-Judge} protocol. We detail the architecture next.

\section{SLM-Based Agentic AI with P–C–G}
We propose an agentic AI architecture built on SLM. The system is decomposed into three modules—\textbf{Planner}, \textbf{Caller}, and \textbf{Generator}—each optimized to perform a distinct role. In this section, we present the overall system design composed of these modules and describe the responsibilities and implementation considerations of each component.

\subsection{System Architecture}
The proposed system consists of three modules:
\vspace{0.5em}
\begin{itemize}
    \item \textbf{Planner}: Analyzes the user's request and produces an execution plan that specifies which tools to call and in what order.
    \item \textbf{Caller}: Instantiates the parameters required by the plan, executes the tools, and collects intermediate results.
    \item \textbf{Generator}: Integrates and reformats the user request and tool outputs to produce the final response aligned with the user's intent.
\end{itemize}
\vspace{0.5em}

In ReAct-style systems that employ SLMs~\cite{yao2023react}, the \textbf{Planner} is repeatedly called for each tool call, such as (P$\rightarrow$C$_1$$\rightarrow$P$\rightarrow$C$_2$$\rightarrow\cdots$). This pattern repeatedly re-processes tool metadata (e.g., \texttt{tool\_desc}) at each Caller step; as the number of tools grows, token usage increases and both system complexity and response latency tend to increase.

\begin{figure}
    \centering
    \includegraphics[width=0.5\textwidth]{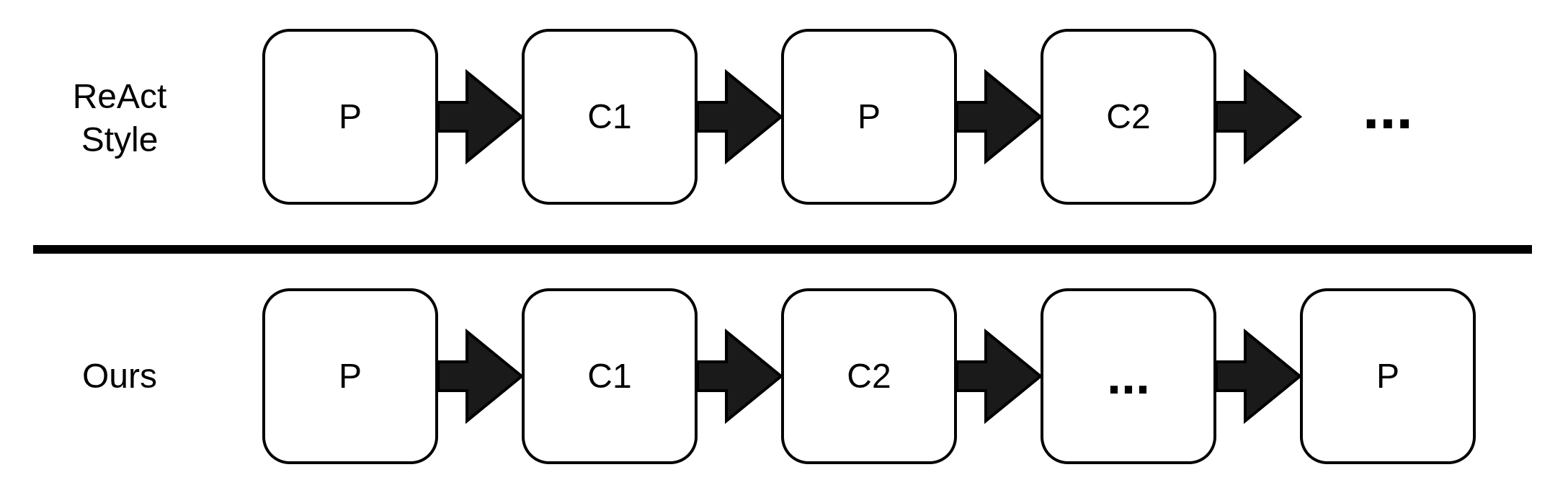}
    \caption{Comparison between a ReAct-style pipeline and the proposed design}
    \label{fig:pcg_restructure}
\end{figure}

By contrast, our design performs a \emph{single upfront plan}: the Planner first drafts the full \texttt{tool\_chain}, the Caller then executes this chain sequentially while aggregating results, and control returns to the Planner only when additional decisions are required (Figure~\ref{fig:pcg_restructure}).

This architectural shift keeps the Planner out of the inner execution loop and invokes it only at decision points, improving efficiency. As a result, it enables a lightweight, SLM-based agentic AI system that reduces computational cost and response latency while maintaining performance on complex tasks.

\begin{figure}
    \centering
    \includegraphics[width=0.5\linewidth]{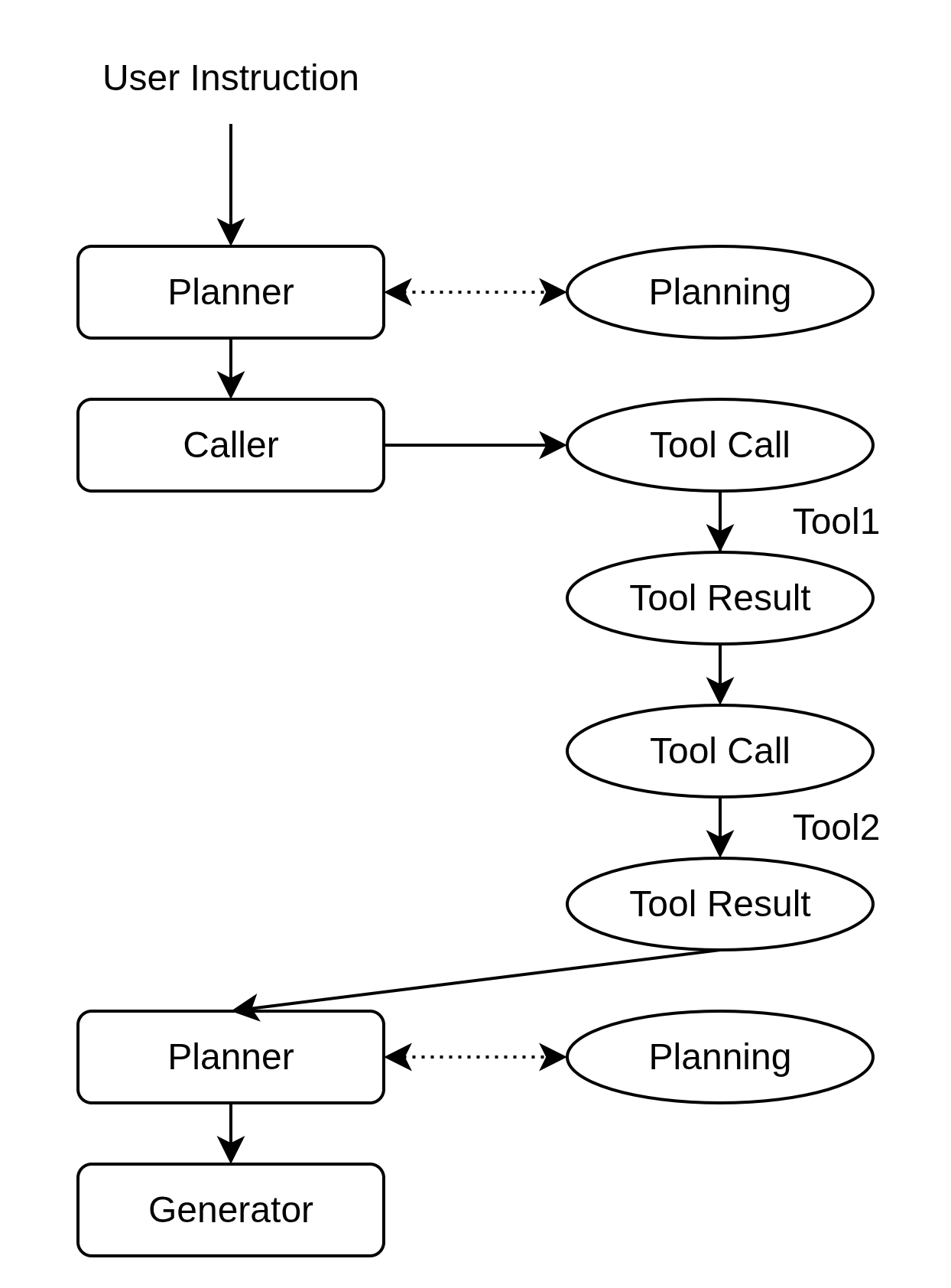}
    \caption{Planner-Caller-Generator Architecture Flow}
    \label{fig:pcg_system_flow}
\end{figure}

The end-to-end dataflow of the Planner–Caller–Generator (P-C-G) system (Figure~\ref{fig:pcg_system_flow}) proceeds as follows:
\vspace{0.5em}
\begin{enumerate}
  \item Upon receiving a user query/instruction, the Planner produces a \texttt{tool\_chain} that specifies the necessary external tools and their call order.
  \item Guided by the \texttt{tool\_chain}, the Caller synthesizes validated parameters for each tool and executes the calls to obtain intermediate results.
  \item Based on the accumulated results, the system decides whether more planning is needed; if information is insufficient, the Planner is re-invoked to update the plan.
  \item Once sufficient information is available, the Generator consolidates the results and renders the final answer in the user-requested format.
\end{enumerate}
\vspace{0.5em}

This workflow enables the P-C-G architecture to handle tasks beyond the capability of a single SLM while maintaining efficiency, by dividing roles into each modules.

\subsection{Planner}
\newlength{\ioheight}
\setlength{\ioheight}{5.4cm}

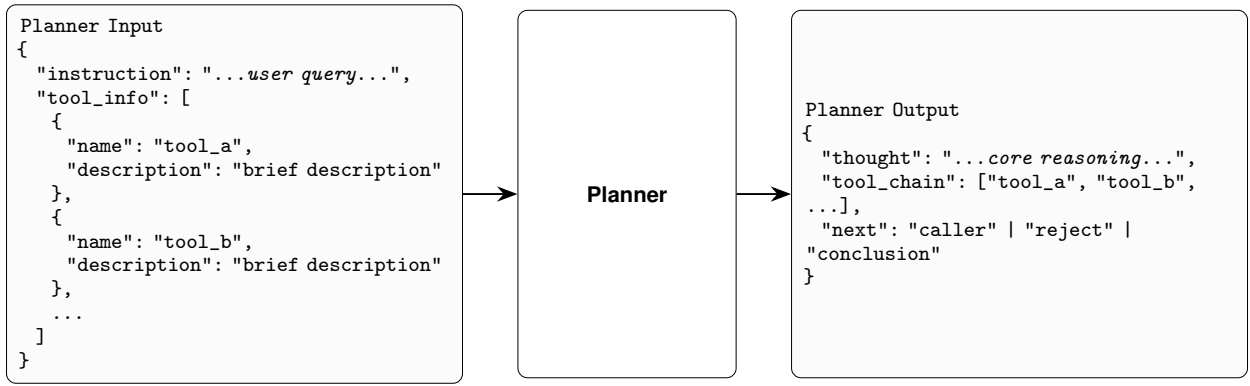
\begin{figure}[t]
  \centering
  \begin{adjustbox}{max width=\linewidth}
  \begin{tikzpicture}[
    every node/.style={transform shape},
    box/.style={draw, rounded corners, inner sep=6pt, align=left, font=\ttfamily\small, fill=gray!3, minimum height=\ioheight},
    proc/.style={draw, rounded corners, inner sep=8pt, align=center, font=\sffamily\small, fill=white, minimum height=\ioheight},
    node distance=8mm
  ]

  \node[box, text width=.38\linewidth] (input) {
    \textbf{Planner Input}\\
    \{\\
    \ \ "instruction": "...\emph{user query}...",\\
    \ \ "tool\_info": [\\
    \ \ \ \ \{ \\
    \ \ \ \ \ \ "name": "tool\_a", \\
    \ \ \ \ \ \ "description": "brief description"\\
    \ \ \ \ \},\\
    \ \ \ \ \{ \\
    \ \ \ \ \ \ "name": "tool\_b", \\
    \ \ \ \ \ \ "description": "brief description"\\
    \ \ \ \ \},\\
    \ \ \ \ \ldots\\
    \ \ ]\\
    \}
  };

  \node[proc, text width=.16\linewidth, right=of input] (planner) {
    \textbf{Planner}
  };

  \node[box, text width=.38\linewidth, right=of planner] (output) {
    \textbf{Planner Output}\\
    \{\\
    \ \ "thought": "...\emph{core reasoning}...",\\
    \ \ "tool\_chain": ["tool\_a", "tool\_b", \ldots],\\
    \ \ "next": "caller" \textbar{} "reject" \textbar{} "conclusion"\\
    \}
  };

  \draw[-{Stealth[length=3mm]}, thick] (input.east) -- (planner.west);
  \draw[-{Stealth[length=3mm]}, thick] (planner.east) -- (output.west);

  \end{tikzpicture}
  \end{adjustbox}
  \caption{The Planner's input/output format}
  \label{fig:planner_io_desc}
\end{figure}

The \textbf{Planner} module analyzes the user's query/instructions to select the necessary external tools and decide the order in which to invoke them, thereby designing the system's overall execution flow. Because a SLM struggles to handle complex problems on its own, the Planner explicitly lays out a solution procedure up front to compensate for the SLM's limitations. It consults the tool list (names and brief descriptions) to quickly determine "what is possible" and constructs an execution path with minimal repetition.

The Planner's input/output format follows Figure~\ref{fig:planner_io_desc}. The input consists of \texttt{instruction} and \texttt{tool\_info}. The \texttt{instruction} contains the body of the user's query/command, and \texttt{tool\_info} lists currently callable tools with their names and short descriptions. Based on this input, the Planner performs internal reasoning and produces an output with \texttt{thought}, \texttt{tool\_chain}, and \texttt{next}. Here, \texttt{thought} summarizes the key judgments and rationales considered to solve the query, \texttt{tool\_chain} specifies the order of tools to invoke, and \texttt{next} is set to one of \texttt{caller}/\texttt{reject}/\texttt{conclusion} to determine the subsequent control flow.

If \texttt{next} is \texttt{caller}, the Caller module generates and validates the input arguments for the next tool in the \texttt{tool\_chain} and executes the call. \texttt{reject} indicates that an immediate answer is not possible due to constraints on available tools/information, in which case the system should clearly communicate the limitation or request additional details from the user. \texttt{conclusion} means sufficient intermediate results have been obtained (or none are needed) to produce a final response, and control is handed off to the Generator. When necessary—specifically when tool execution fails, returns unexpected formats, or produces results that contradict the planned sequence—the system triggers limited replanning by re-invoking the Planner to update the \texttt{tool\_chain} with the new information.

In summary, the Planner aligns user intent with available tool capabilities to produce an executable plan and to decide control flow to Caller/Generator, thereby supporting the efficiency and robustness of the overall P–C–G pipeline.

\subsection{Caller}
\begin{figure}[t]
  \centering
  \begin{adjustbox}{max width=.86\linewidth} 
  \begin{tikzpicture}[
    every node/.style={transform shape},
    box/.style={draw, rounded corners, inner sep=6pt, align=left,
                font=\ttfamily\footnotesize, fill=gray!3,
                text width=12.5cm, minimum height=\ioheight},
    proc/.style={draw, rounded corners, inner sep=8pt, align=center,
                 font=\sffamily\small, fill=white,
                 text width=3cm, minimum height=1.1cm},
    node distance=6mm
  ]

  \node[box] (input) {
    \textbf{Caller Input}\\
    \{\\
    \ \ "messages":[...],\\
    \ \ "tools": \\
    \ \ \ \{\\
    \ \ \ \ "name":"<tool\_name>",\\
    \ \ \ \ "desc":"tool description",\\
    \ \ \ \ "parameters": \\
    \ \ \ \ \ \ \{ \\
    \ \ \ \ \ \ \ "type":"object", \\
    \ \ \ \ \ \ \ "properties": [\\
    \ \ \ \ \ \ \{\\
    \ \ \ \ \ \ \ \ "<param1>":\{ \\
    \ \ \ \ \ \ \ \ "type":"string", \\
    \ \ \ \ \ \ \ \ "description":"param1 description" \\
    \ \ \ \ \ \ \},\\
    \ \ \ \ \ \ \{ "<param2>":\{ "type":"\ldots", "description":"\ldots"\}\},\\
    \ \ \ \ \ \ \ldots \\
    \ \ \ \ \ \ ], \\
    \ \ \ \ \ "required":["<required param>", \ldots] \\
    \ \ \ \ \ \} \\
    \ \ \ \} \\
    \}
  };

  \node[proc, below=6mm of input] (caller) {\textbf{Caller}};

  \node[box, below=6mm of caller] (output) {
     \textbf{Caller Output}\\
      \{\\
      \ \ "name": "<tool\_name>",\\
      \ \ "arguments": \{ \\
      \ \ \ \ "<param1>": "param1 value",\\
      \ \ \ \ "<param2>": "param2 value",\\
      \ \ \ \ \ldots \\
      \ \ \} \\
      \}
  };

  \draw[-{Stealth[length=3mm]}, thick] (input.south) -- (caller.north);
  \draw[-{Stealth[length=3mm]}, thick] (caller.south) -- (output.north);

  \end{tikzpicture}
  \end{adjustbox}
  \caption{The Caller's input/output format}
  \label{fig:caller_io_desc}
\end{figure}
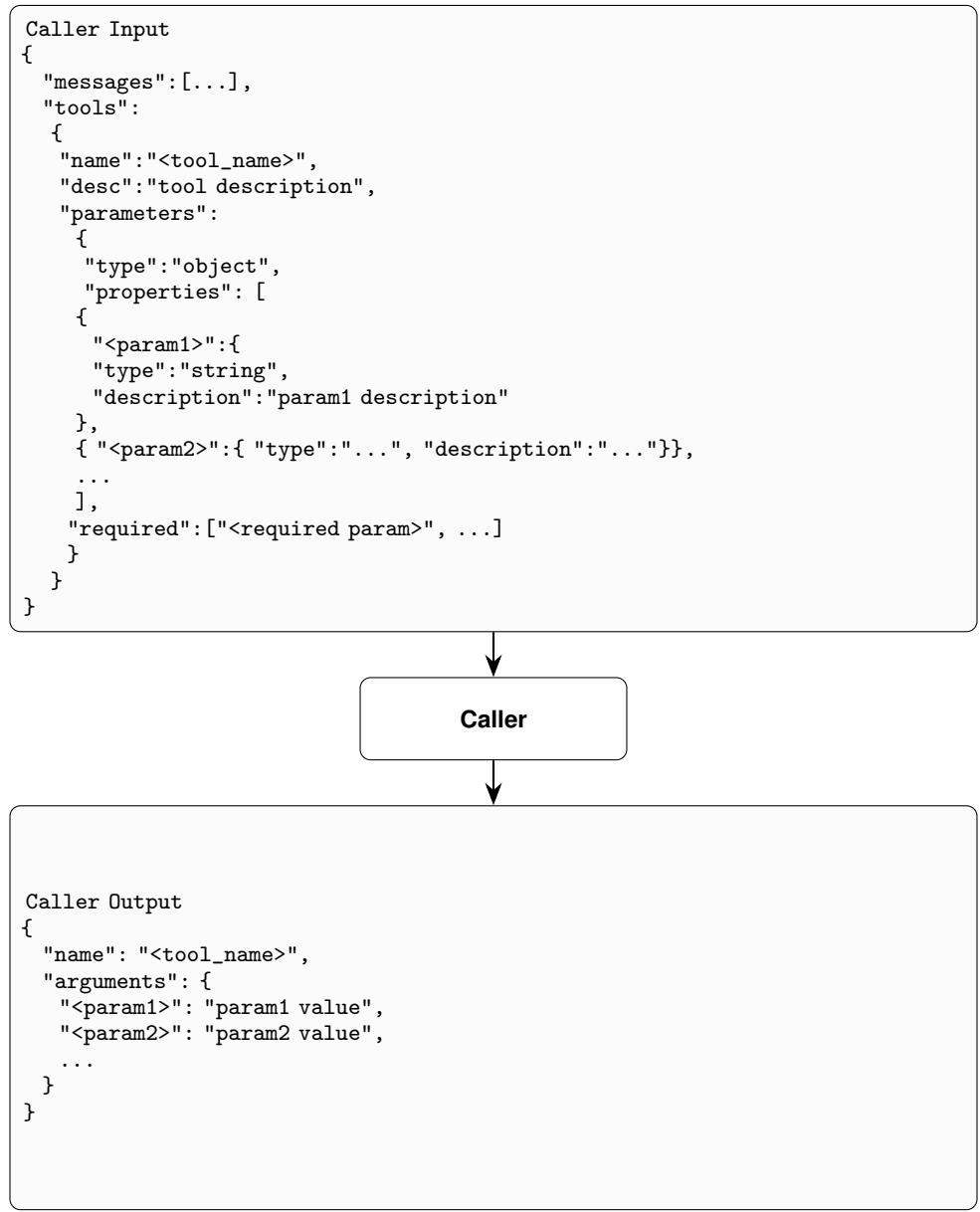

The \textbf{Caller} module refines the Planner's plan into an executable call specification and, when needed, executes the tool and prepares outputs for next steps. Its workflow consists of (i) tool-call preparation (argument synthesis and schema validation), (ii) tool execution, and (iii) result normalization and next-step decision. To ensure reliability in Korean settings, we adopt a \emph{Korean-first value policy}: value fields are kept in Korean by default, with whitelist exceptions only for fields whose schemas permit English-only input. The synthesized arguments undergo pre-/post-validation for required vs. optional fields, types, and admissible ranges (e.g., enums, patterns, min/max).

For a simplified external interface, the Caller's default output is always the following standardized call object: \\
\begin{verbatim}
{
"name": "<tool_name>",
"arguments": {
"<param1>": "<param1 value>",
... }
}
\end{verbatim}

The Caller strictly follows the \texttt{tool\_chain} fixed by the Planner and, at each step, packages the target tool's \texttt{"name"} and its \texttt{"arguments"} into this standardized object. Once preparation is complete, it performs the actual call using this object, normalizes the tool's response to a common schema, and performs error checks. If no issues are found, control proceeds to the next step in the chain; otherwise, the module handles the condition (e.g., missing required parameters or malformed/inconsistent arguments) and determines the subsequent flow.

\subsection{Generator}
\begin{figure}[t]
  \centering
  \includegraphics[width=.85\linewidth]{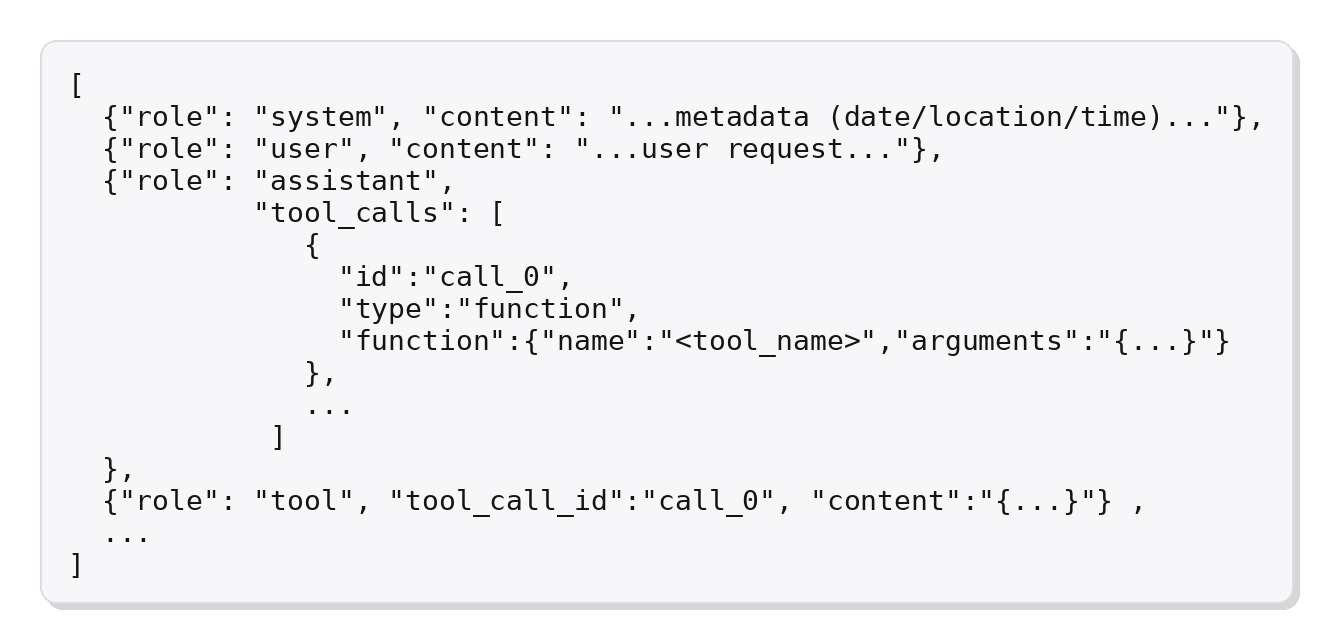}
  \caption{Minimal tool-use message structure}
  \label{fig:functionchat_msg}
\end{figure}

The \textbf{Generator} module integrates and normalizes the tool outputs collected across the Planner–Caller stages, then interprets, summarizes, and restructures them to produce a final response that matches the user's intent and desired format. In this work, the Generator uses the base model without additional fine-tuning and operates with a minimal \texttt{chat\_template}. A compact view that includes only the fields essential for generation is shown in Figure~\ref{fig:functionchat_msg}.

The \textbf{system} turn carries lightweight metadata (e.g., date and location), the \textbf{user} turn contains the user's request, the \textbf{assistant} turn records only the tool-call specification (\texttt{tool\_calls}), and the \textbf{tool} turn returns the execution result linked via \texttt{tool\_call\_id}. Leveraging these standardized tool results together with the dialogue context, the Generator avoids mere enumeration and produces the final natural-language answer, applying appropriate formats such as tables, summaries, or reports.

\section{Experiments}
\subsection{Evaluation Dataset}
We constructed a Korean tool-use evaluation dataset to closely mirror real service conditions. In contrast to prior evaluations that focused primarily on plain response generation, our dataset reflects composite scenarios involving tool utilization, information completion, and constraint awareness. It comprises four categories, which are designed to be evenly distributed across types.

\begin{itemize}
  \item \textbf{Single-chain}: Evaluates whether a model can answer a user query with a \emph{single} tool call. These queries do not depend on prior dialogue context and typically involve simple Q\&A, data lookup, or lightweight content generation.
  \item \textbf{Multi-chain}: Evaluates a model's ability to process a request by invoking \emph{multiple} tools in sequence. The model must determine a correct call order and manage inter-tool dependencies to produce a complete and accurate final answer.
  \item \textbf{Missing Parameters}: Evaluates recognition of missing \emph{required} parameters in single-tool scenarios where the absent values cannot be inferred from internal context. The model should avoid unfounded guesses, identify the missing fields, and request the user to provide the necessary information.
  \item \textbf{Missing Functions}: Evaluates whether the model can detect that the \emph{necessary} tool is not present in the provided tool list. Upon recognizing this constraint, the model should clearly inform the user and, when possible, suggest alternative paths to achieve the same goal.
\end{itemize}

The evaluation dataset contains \textbf{400} instances in total, with \textbf{100} items per category. We curated tool definitions from public API specifications, then reproduced realistic user queries and generated the data via a hybrid automatic–manual process followed by human verification. Each example includes (i) a user query, (ii) the available tool specification, and (iii) the tool-call result.

\begin{figure}[t]
  \centering
  \begin{subfigure}{.49\linewidth}
    \centering
    \includegraphics[width=\linewidth]{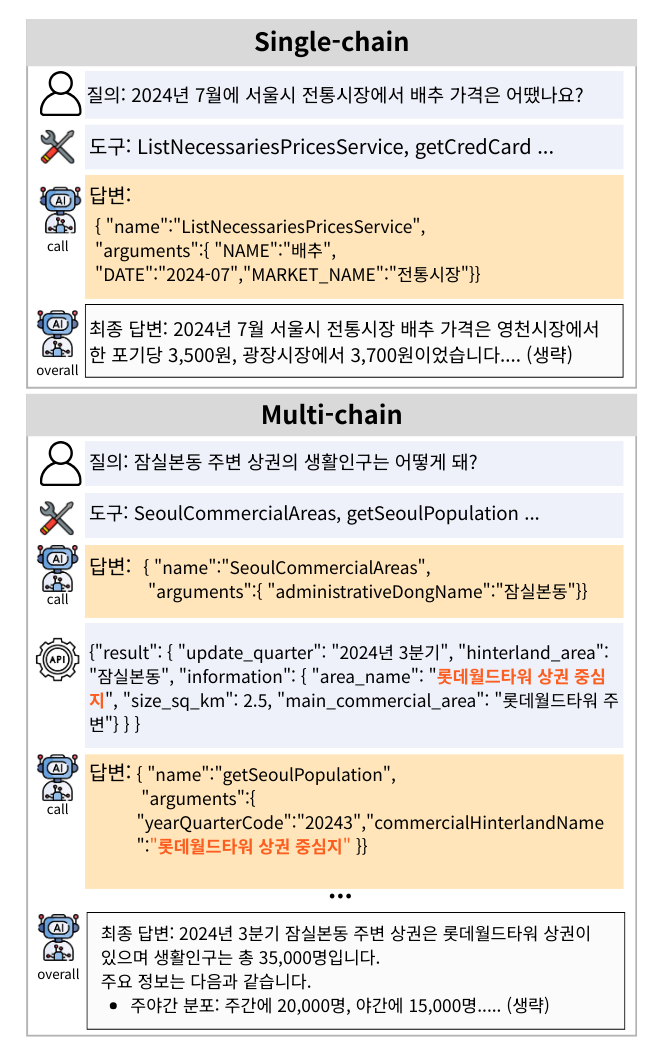}
    \caption{Korean version}
    \label{fig:single_multi_chain}
  \end{subfigure}
  \hfill
  \begin{subfigure}{.49\linewidth}
    \centering
    \includegraphics[width=\linewidth]{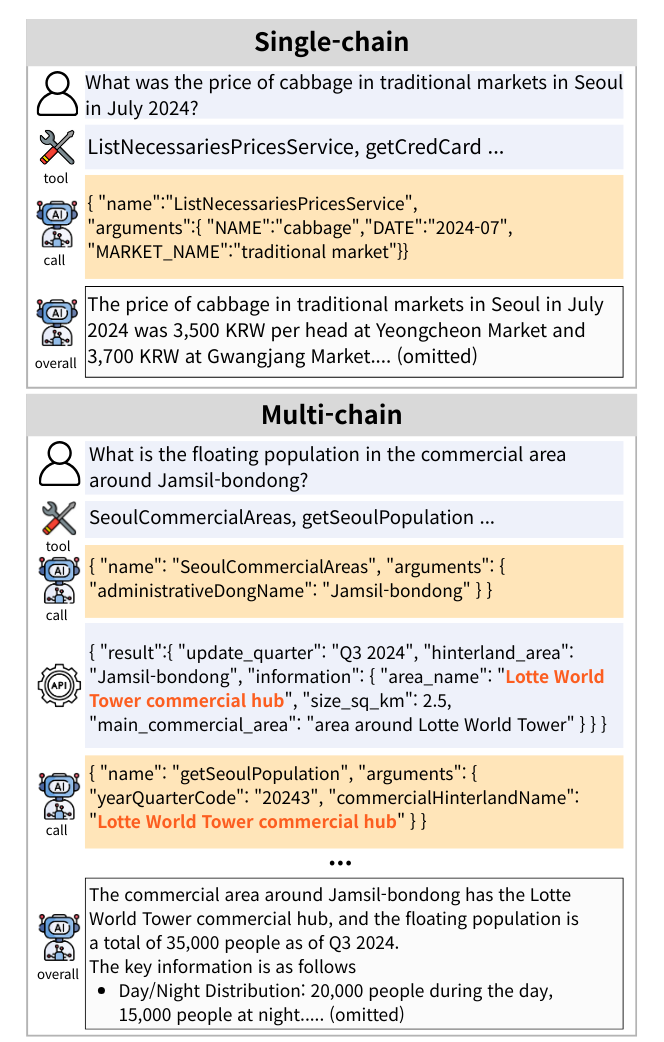}
    \caption{English counterpart (for illustration only)}
    \label{fig:single_multi_chain_en}
  \end{subfigure}

  \vspace{2pt}
  \caption{Single-/Multi-chain evaluation example in Korean (left) and its English counterpart (right)}
  \label{fig:single_multi_chain_both}
\end{figure}

In the multi-chain example(Figure \ref{fig:single_multi_chain}) of the evaluation dataset, the output of the first call (e.g., "Lotte World Tower commercial hub") is passed as the input to the subsequent call, enabling us to test the model's ability to track dialogue history and manage inter-tool dependencies. This setting evaluates the stepwise reasoning and tool-to-tool coordination required in real service environments. We additionally include an English counterpart of this example (Fig.~\ref{fig:single_multi_chain_en}) that mirrors the semantics of the Korean version. This figure is provided for illustration only and was not used for model training or evaluation.



\subsection{Experimental Setup and Metrics}
Our experiments are designed to evaluate both the individual modules—\textbf{Planner}, \textbf{Caller} and \textbf{Generator}—and the overall P–C–G architecture within a single framework. We consider two complementary settings that differ in API usage and task composition.

\paragraph{Setting I — End-to-End (Planner / Overall Architecture)}
We run \textbf{live API calls} on a total of 300 tasks spanning Single-chain, Multi-chain, and Missing Functions. By passing each model's generated arguments directly to the APIs and obtaining real responses, we jointly assess \emph{planning quality}, \emph{Task Success Rate}, and \emph{inference efficiency (tokens/latency)} (reported metrics: Planner, TSR, Tokens/Latency). To fairly reflect differences in argument generation across models, we adopt live (non-deterministic) API responses rather than fixed outputs. Missing Parameters is excluded here because it does not directly probe Planner quality.\footnote{In Missing Parameters, the correct action often reduces to "do not call," which makes it difficult to separate planning quality and tends to overlap with the Caller-focused evaluation in Setting II.}

\paragraph{Setting II — Module-level (Caller / Generator)}
To fairly compare \emph{argument-generation accuracy} and \emph{final response quality} for single-tool calls, we prepare \textbf{deterministic API responses} and feed identical inputs to all models (reported metrics: Planning Performance; Tool-Use Evaluation; Execution-Stage Evaluation; Task Success Rate; Inference Efficiency (tokens/latency)). This setting removes external API variability and focuses on measuring the Caller's schema adherence and argument completeness, as well as the Generator's factuality, structure, and tone.

All evaluations assume Korean user queries with Korean tool/parameter specifications, and models are required to preserve Korean in outputs. Scores are computed via an \emph{LLM-as-a-Judge} protocol using \textit{Claude 4} averaged over five runs.

\paragraph{(A) Planning Evaluation} 
Because the quality of the initial plan directly affects token usage and response latency, we evaluate the Planner’s planning accuracy separately. We first restrict attention to cases judged correct by an \emph{LLM-as-a-Judge}; for those cases only, we compare the actual number of tool calls with the human-defined optimal number.

In the P–C–G architecture we propose, the Planner’s pre-generated \texttt{tool\_chain} is treated as the initial plan. By contrast, the comparison models (GPT-4o-mini, EXAONE-4.0-32B-AWQ, Qwen3-14B, Qwen-8B) do not have a separate Planner; for them, the number of tool calls actually invoked during tool use is taken as their "planned" call count. This evaluation therefore contrasts (i) the Planner’s proposed initial plan and (ii) the baselines’ realized call counts against the human reference chain(\(k_{\text{opt}}\)) to test whether better initial planning improves efficiency.

Let \(\mathcal{C}\) be the set of samples judged correct. For each sample \(i\), let \(k_{\hat{i}}\) denote the actual number of tool calls, and \(k_{\text{opt},i}\) the optimal number from the reference chain. We define the planning-quality metrics as:

\[
\begin{aligned}
\textbf{Under\mbox{-}planning} \; &= 
\frac{\left|\{\, i\in\mathcal{C}\mid k_{\hat{i}}<k_{\text{opt},i}\,\}\right|}{\left|\mathcal{C}\right|},\\[4pt]
\textbf{As\mbox{-}planned} \; &= 
\frac{\left|\{\, i\in\mathcal{C}\mid k_{\hat{i}}=k_{\text{opt},i}\,\}\right|}{\left|\mathcal{C}\right|},\\[4pt]
\textbf{Over\mbox{-}planning} \; &= 
\frac{\left|\{\, i\in\mathcal{C}\mid k_{\hat{i}}>k_{\text{opt},i}\,\}\right|}{\left|\mathcal{C}\right|}.
\end{aligned}
\]

\textit{Under-planning} indicates fewer calls than optimal (risk of omission), \textit{As-planned} indicates exact agreement with the optimal plan (parsimonious, adequate calls), and \textit{Over-planning} indicates unnecessary extra calls (token/latency overhead).

\paragraph{(B) Tool-Use Evaluation}
We comprehensively assess the model’s ability to execute from single tool calls up to 2–4 step chained calls. A step-wise single-chain and multi-chain dataset is used, and accuracy is measured at each step.

\begin{itemize}
\item \textbf{Step-wise Call Accuracy} \
Evaluation of correct tool selection and accurate argument generation given the tool schema. In the $n$-step setting, we provide ground-truth calls and tool responses up to step 
$(n\!-\!1)$ and measure the accuracy of the generated call arguments at step $n$.
\end{itemize}

\paragraph{(C) Execution-Stage Evaluation (Caller/Generator)}
Quality of tool invocation and final answer generation is measured with the following metrics.
\begin{itemize}
    \item \textbf{Single-chain / Multi-chain}
    \begin{itemize}
        \item \emph{Overall}: Factuality, structure, and tone of the final response
        \item \emph{Call Accuracy}: Completeness and correctness of tool selection and argument generation
    \end{itemize}
    \item \textbf{Constraint Awareness}
    \begin{itemize}
        \item \emph{Missing Parameters / Missing Functions}: Appropriateness of diagnosis, avoidance of unfounded guesses, and quality of recovery behavior (e.g., asking the user for missing information or proposing alternatives)
    \end{itemize}
\end{itemize}

\paragraph{(D) Task Success Rate Evaluation}
\begin{itemize}
    \item \textbf{Task Success Rate(TSR)}: A task is judged successful if and only if the model’s final answer satisfies all three criteria: (i) it addresses the query’s core intent, (ii) it uses tools in a logically sound manner, and (iii) it is complete (i.e., not missing required information). Decisions are made by an \emph{LLM-as-a-Judge}; minor timing differences from API calls or superficial formatting variations do not affect the verdict. TSR is defined as the proportion of evaluated queries judged successful.
\end{itemize}

\paragraph{(E) Inference Efficiency (Tokens/Latency)}
We evaluate how P–C–G affects \textbf{token usage} and \textbf{response latency} during inference.

\begin{itemize}
  \item \textbf{Tokens Average}: For queries correctly answered by all models, the total number of tokens used to process a single query
  \item \textbf{Latency}: For queries correctly answered by all models, the total time taken to produce the final answer.
\end{itemize}

\subsection{Experimental Results}
\subsubsection{Planning}
According to Table~\ref{tab:over-planner}, the proposed architecture attains the highest \textbf{As-planned} rate (92.3\%) and a low \textbf{Over-planning} rate (1.6\%), confirming the effectiveness of the \emph{initial single plan + limited on-demand replanning} strategy. By contrast, GPT-4o-mini and EXAONE-4.0-32B-AWQ exhibit \emph{Over-planning} rates above 6\%, indicating a higher risk of unnecessary token usage and latency.

\begin{table}[h]
\centering
\small
\begin{tabular}{l c c c}
\toprule
Model  & Under-planning (\%) & As-planned (\%) & Over-planning (\%) \\
\midrule
GPT-4o-mini & 4.3 & 89.6 & 6.1 \\
EXAONE-4.0-32B-AWQ  & 4.4 & 89.6 & 6.0 \\
Qwen3-14B    & 8.3 & 89.5 & 2.2 \\
Qwen3-8B     & 7.9 & 90.9 & 1.2 \\
Ours        & \textbf{6.1} & \textbf{92.3} & \textbf{1.6} \\
\bottomrule
\end{tabular}
\vspace{0.5em}
\caption{Planning Performance by Model (\%)}
\label{tab:over-planner}
\end{table}

\subsubsection{Tool Use}
According to Table~\ref{tab:tool-call}, the Caller module of the proposed architecture attains a tool-use accuracy of 75.0\%, outperforming all compared models, including GPT-4o-mini (70.5\%), EXAONE-4.0-32B-AWQ (71.4\%), Qwen3-14B (68.6\%), and Qwen3-8B (59.6\%).

\begin{table}[h!]
\centering
\begin{tabular}{l c}
\toprule
Model & Call Accuracy (\%) \\
\midrule
GPT-4o-mini & 70.5 \\
EXAONE-4.0-32B-AWQ & 71.4 \\
Qwen3-14B & 68.6 \\
Qwen3-8B & 59.6 \\
Ours & \textbf{75.0} \\
\bottomrule
\end{tabular}
\vspace{0.5em}
\caption{Comparison of Tool-Use Performance by Model (\%)}
\label{tab:tool-call}
\end{table}

\subsubsection{Single-chain}
In the Single-chain tasks (Tables~\ref{tab:single-call}, \ref{tab:single-overall}), P–C–G achieves a \textbf{Call Accuracy} of 95.6\% and an \textbf{Overall} score of 91.2\%, outperforming Qwen3-14B (93.0\% / 88.4\%) and GPT-4o-mini (86.0\% / 87.4\%).

\begin{table}[h!]
\centering
\begin{tabular}{l c}
\toprule
Model &  Call Accuracy (\%)\\
\midrule
GPT-4o-mini & 86.0 \\
EXAONE-4.0-32B-AWQ & 88.6 \\
Qwen3-14B & 93.0 \\
Qwen3-8B & 90.6 \\
Ours & \textbf{95.6} \\
\bottomrule
\end{tabular}
\vspace{0.5em}
\caption{Comparison of Single-chain Performance by Model (\%)}
\label{tab:single-call}
\end{table}

\begin{table}[h!]
\centering
\begin{tabular}{l c}
\toprule
Model & Overall Accuracy (\%) \\
\midrule
GPT-4o-mini & 87.4 \\
EXAONE-4.0-32B-AWQ & 80.2 \\
Qwen3-14B & 88.4 \\
Qwen3-8B & 87.8 \\
Ours & \textbf{91.2} \\
\bottomrule
\end{tabular}
\vspace{0.5em}
\caption{Comparison of Single-chain Overall Performance by model (\%)}
\label{tab:single-overall}
\end{table}

\subsubsection{Multi-chain}
In the Multi-chain tasks (Tables~\ref{tab:multi-call}, \ref{tab:multi-overall}), the proposed architecture achieves a \textbf{Call Accuracy} of 33.8\%, slightly surpassing GPT-4o-mini (33.0\%; +0.8 percentage points) and clearly outperforming EXAONE-4.0-32B-AWQ (32.6\%) and Qwen3-14B (28.4\%). For \textbf{Overall}, GPT-4o-mini ranks first at 67.0\%, while our system places second at 62.4\%, well ahead of EXAONE-4.0-32B-AWQ (56.6\%) and the Qwen3 family.

\begin{table}[h!]
\centering
\begin{tabular}{l c}
\toprule
Model & Call Accuracy (\%) \\
\midrule
GPT-4o-mini & 33.0 \\
EXAONE-4.0-32B-AWQ & 32.6 \\
Qwen3-14B & 28.4 \\
Qwen3-8B & 20.4 \\
Ours & \textbf{33.8} \\
\bottomrule
\end{tabular}
\vspace{0.5em}
\caption{Comparison of Multi-chain Performance by Model (\%)}
\label{tab:multi-call}
\end{table}

\begin{table}[h!]
\centering
\begin{tabular}{l c}
\toprule
Model & Overall Accuracy (\%) \\
\midrule
GPT-4o-mini & \textbf{67.0} \\
EXAONE-4.0-32B-AWQ & 56.6 \\
Qwen3-14B & 58.2 \\
Qwen3-8B & 47.0 \\
Ours & 62.4 \\
\bottomrule
\end{tabular}
\vspace{0.5em}
\caption{Comparison of Multi-chain Overall Performance by Model (\%)}
\label{tab:multi-overall}
\end{table}

\subsubsection{Missing Parameters}
According to Table~\ref{tab:missing-parameter}, the proposed architecture achieves 66.6\% on Missing Parameters—slightly below GPT-4o-mini (76.8\%) and Qwen3-14B (76.8\%)—but clearly ahead of EXAONE-4.0-32B-AWQ (53.6\%) and Qwen3-8B (56.8\%).

\begin{table}[h!]
\centering
\begin{tabular}{l c}
\toprule
Model & Accuracy (\%) \\
\midrule
GPT-4o-mini & 76.8 \\
EXAONE-4.0-32B-AWQ & 53.6 \\
Qwen3-14B & 76.8 \\
Qwen3-8B & 56.8 \\
Ours & \textbf{66.6} \\
\bottomrule
\end{tabular}
\vspace{0.5em}
\caption{Comparison of Missing Parameters Performance by Model (\%)}
\label{tab:missing-parameter}
\end{table}

\subsubsection{Missing Functions}
In \textbf{Missing Functions} (Table~\ref{tab:missing-function}), the proposed architecture attains \textbf{91.2\%}, the highest among all models—surpassing Qwen3-14B (90.0\%) and Qwen3-8B (87.2\%), and by a wide margin over EXAONE-4.0-32B-AWQ (75.2\%) and GPT-4o-mini (65.0\%).

\begin{table}[h!]
\centering
\begin{tabular}{l c}
\toprule
Model & Accuracy (\%) \\
\midrule
GPT-4o-mini & 65.0 \\
EXAONE-4.0-32B-AWQ & 75.2\\
Qwen3-14B & 90.0  \\
Qwen3-8B & 87.2\\
Ours& \textbf{91.2} \\
\bottomrule
\end{tabular}
\vspace{0.5em}
\caption{Comparison of Missing Functions Performance by Model (\%)}
\label{tab:missing-function}
\end{table}

\subsubsection{Task Success Rate}
According to Table~\ref{tab:Task Success Rate}, the proposed architecture achieves the highest TSR at \textbf{79.7\%}. While the +0.4 percentage-point improvement over GPT-4o-mini (79.3\%) and EXAONE-4.0-32B-AWQ (78.6\%) is not statistically significant, it remains competitive despite relying only on an SLM. It also surpasses the performance of the same base model, Qwen-8B (72.4\%). These results suggest that the P–C–G design is effective with respect to task success.

\begin{table}[h]
\centering
\small
\begin{tabular}{l c}
\toprule
Model & TSR (\%)  \\
\midrule
GPT-4o-mini& 79.3  \\
EXAONE-4.0-32B-AWQ  & 78.6 \\
Qwen3-14B   & 76.9  \\
Qwen3-8B    & 72.4 \\
Ours        & \textbf{79.7} \\
\bottomrule
\end{tabular}
\vspace{0.5em}
\caption{Comparison of Task Success Rate Results by Model (\%)}

\label{tab:Task Success Rate}
\end{table}

\subsubsection{Inference Efficiency (Tokens/Latency)}
According to Table~\ref{tab:inference_efficiency}, the P–C–G architecture uses an average of 4,360.3 tokens—the fewest among SLM—while achieving a high accuracy of 79.7\%. This corresponds to a token reduction of approximately 12–22\% compared to other models.

Notably, EXAONE-4.0-32B-AWQ’s over-planning led to unnecessary re-calls and increased token usage (5,337.8 tokens), whereas P–C–G mitigated such inefficiency by establishing an accurate initial plan.

In terms of inference time, P–C–G maintains a competitive 9.1 seconds, securing both the compactness and efficiency advantages of an SLM-based design. These results empirically demonstrate that an "\emph{initial single plan + limited on-demand replanning}" strategy can improve quality while simultaneously enhancing token efficiency.

\begin{table}[h]
\centering
\small
\begin{tabular}{lcc}
\toprule
Model & Tokens Average & Latency (s) \\
\midrule
GPT-4o-mini & 3747.6 & 7.9 \\
EXAONE-4.0-32B-AWQ & 5337.8 & 9.6 \\
Qwen3-14B  & 4980.6 & 7.5 \\
Qwen-8B   & 4950.3 & \textbf{6.1} \\
Ours      & \textbf{4360.3} & 9.1 \\
\bottomrule
\end{tabular}
\vspace{0.5em}
\caption{Average token usage and response time by model}
\label{tab:inference_efficiency}
\end{table}

\subsection{Discussion}
The proposed P-C-G architecture achieves both efficiency and stability through an "\emph{initial single plan + limited on-demand replanning}" strategy. Below, we summarize the main findings in the order of the results section (Planning → Tool Calling → Single-chain → Multi-chain → Constraints → TSR → Efficiency).

\textbf{(1) Planning Performance} According to Table~\ref{tab:over-planner}, P-C-G shows the highest \textbf{As-planned} rate at \textbf{92.3\%} among the correct-answer cases, and a low \textbf{Over-planning} rate at \textbf{1.6\%}. In contrast, EXAONE-4.0-32B-AWQ has an \textbf{Over-planning} rate of 6.0\%, indicating a relatively large number of unnecessary calls. This demonstrates that the accuracy of the initial plan directly affects subsequent inference cost and quality.

\textbf{(2) Tool Calling} According to Table~\ref{tab:tool-call}, the Caller module’s \textbf{Call Accuracy} is \textbf{75.0\%}, outperforming all comparison models. This can be attributed to P-C-G’s role-separated design, which streamlines tool selection and parameter generation.

\textbf{(3) Single-chain} On single-chain tasks, it achieved top performance in both \textbf{Call} \textbf{95.6\%} and \textbf{Overall} \textbf{91.2\%} (Tables~\ref{tab:single-call} and \ref{tab:single-overall}). This is attributable to the Planner yielding a minimal call path without unnecessary branching for simple requests, and the Caller/Generator executing it without degradation.

\textbf{(4) Multi-chain} On multi-chain tasks, it achieved the highest \textbf{Call} at \textbf{33.8\%}, and the second-highest \textbf{Overall} at \textbf{62.4\%} (Tables~\ref{tab:multi-call}, \ref{tab:multi-overall}). Although GPT-4o-mini led in \textbf{Overall} with 67.0\%, P-C-G maintained relatively stable accuracy in tool selection and argument generation across the entire chain. There is room for further improvement by strengthening multi-step state management and consistency checks of intermediate results.

\textbf{(5) Constraint Awareness} In \textbf{Missing Parameters}, it scored \textbf{66.6\%}, lower than some larger models (Table~\ref{tab:missing-parameter}), but in \textbf{Missing Functions} it achieved \textbf{91.2\%}, the best score (Table~\ref{tab:missing-function}). In other words, while fine-grained value completion—such as inferring missing arguments—remains a limitation of SLMs, its ability to identify and block tools not present in the provided list at an early stage is strong.

\textbf{(6) Task Success Rate} The \textbf{Task Success Rate} is \textbf{79.7\%}, on par with GPT-4o-mini (79.3\%) (Table~\ref{tab:Task Success Rate}) and clearly higher than Qwen-8B (72.4\%). From a TSR perspective, the absolute gap with large models is small, but achieving comparable performance with a lightweight model carries significant cost–performance implications.

\textbf{(7) Inference Efficiency} In terms of tokens and latency, P-C-G used an average of \textbf{4360.3 tokens} for correct answers—the fewest among SLM (a 12–22\% reduction)—and maintained a practical response time of \textbf{9.1 s} (Table~\ref{tab:inference_efficiency}). Although the Planner stage adds an average of 2.8 seconds, the accurate \texttt{tool\_chain} reduces the Caller’s input tokens, offsetting and improving overall efficiency. Further Planner lightweighting could reduce latency even more.

Meanwhile, despite a relatively high Over-planning rate (6.1\%), GPT-4o-mini recorded the lowest average token count (3747.6) and the shortest latency (7.9s). This contrasts with other models (e.g., EXAONE-4.0-32B-AWQ), where Over-planning generally leads to increased token usage and response delays, and it is likely related to the \texttt{Prompt\_caching} feature applicable when using the OpenAI function-calling API. In the same context, when calls are repeatedly invoked within the same context, portions of the input prompt can be cached, reducing the number of tokens actually billed/processed and thereby shortening response time. Therefore, GPT-4o-mini’s efficiency figures may have been influenced by this caching mechanism, and caution is warranted when directly comparing it with other models that do not apply caching. \footnote{\url{https://cookbook.openai.com/examples/prompt_caching101}}

\textbf{(8) Comprehensive Analysis} EXAONE-4.0-32B-AWQ is competitive in Tool Calling but tends to suffer reduced efficiency on simple tasks due to excessive planning and re-calls. GPT-4o-mini shows strength in Multi-chain \textbf{Overall}, but on simple tasks its efficiency drops due to over-reasoning. Qwen3 family shows robust performance in single-chain and constraint awareness, but shows weaknesses in multi-chain and state management. In contrast, despite being SLM-based, the proposed architecture achieves balanced performance and efficiency across all categories, confirming that its design simultaneously satisfies real-world demands for "\textbf{accuracy + cost efficiency}".
 
\section{Conclusion}
This study proposes a \textbf{Planner–Caller–Generator (P–C–G)} architecture that combines the lightweight nature of SLMs with agentic techniques, and demonstrates its utility through an evaluation optimized for Korean tool-use environments. Through \emph{initial planning + limited on-demand replanning}, a \emph{Korean-first value policy}, and a Caller module that uses \emph{normalized tool-call objects} with simultaneous schema/value validation, we achieve reliable tool calls and response generation with a small parameter size. The evaluation protocol assumes Korean queries, tools, and parameter specifications; it covers Single/Multi-chain and Missing Parameters/Functions, and secures fairness via an \emph{LLM-as-a-Judge} protocol averaged over five runs.

Across experiments, P–C–G simultaneously achieved high initial planning accuracy and suppression of unnecessary calls, showing strong tool selection and argument generation in both single and multi-chain tool calls. Especially, it excelled at early identification and safe handling of missing-function situations, while maintaining practical performance even on fine-grained reasoning tasks such as missing-argument completion. Leveraging its lightweight design, it delivered cost-efficient inference without quality degradation in terms of token usage and latency.

We have demonstrated that even with SLM-based agentic AI, tool-use accuracy, predictability, and cost-effectiveness can be achieved simultaneously through role separation and interface standardization. However, the Planner’s inference step introduces a modest latency overhead, and the missing parameter task remains an area for improvement, compared with larger models.

Future work includes: (1) reducing latency via Planner lightweighting/caching, (2) improving parameter completion, (3) uncertainty-aware replanning for partial-failure recovery, and (4) defenses for security/privacy, including protection against prompt injection and data exfiltration. Overall, this study demonstrates the practical viability of SLM-based agentic AI in Korean-centric, real-world settings and suggests it as a meaningful alternative where cost–performance is paramount.

\bibliographystyle{unsrt}  
\bibliography{references}  

\end{document}